\pdfoutput=1
\documentclass[conference]{IEEEtran}
\PassOptionsToPackage{numbers,square,sort&compress}{natbib}
\usepackage{times}

\usepackage[numbers]{natbib}
\usepackage{multicol}
\usepackage[bookmarks=true,colorlinks=true,urlcolor=black]{hyperref}

\usepackage{graphicx}
\usepackage{amsmath}
\usepackage{amssymb}
\usepackage{booktabs}
\usepackage{cuted}
\usepackage{amsmath}
\usepackage{tocloft}
\usepackage{etoc} %

\usepackage[capitalize]{cleveref}
\Crefname{section}{Sec.}{Secs.}
\crefname{section}{Sec.}{Secs.}
\Crefname{table}{Tab.}{Tabs.}
\crefname{table}{Tab.}{Tabs.}
\Crefname{figure}{Fig.}{Figs.}
\crefname{figure}{Fig.}{Figs.}
\Crefname{appendix}{App.}{Apps.}
\crefname{appendix}{App.}{Apps.}

\usepackage[utf8]{inputenc} %
\usepackage[T1]{fontenc}    %
\usepackage{url}            %
\usepackage{amsfonts}       %
\usepackage{nicefrac}       %
\usepackage[nopatch=eqnum]{microtype}      %

\usepackage{epsfig}
\usepackage{tabularx}
\usepackage{listings}
\usepackage{color}
\definecolor{physTwinCitationBlue}{RGB}{48,131,205}
\hypersetup{citecolor=physTwinCitationBlue,linkcolor=physTwinCitationBlue}
\usepackage[most]{tcolorbox} %
\usepackage[framemethod=default]{mdframed} %
\usepackage{wrapfig}
\usepackage{subcaption}
\usepackage{float}
\usepackage{makecell}
\usepackage[percent]{overpic}
\usepackage{adjustbox}

\usepackage{multirow}
\usepackage{comment}
\usepackage{xspace}

\newcommand{\stdhide}[1]{}

\newcolumntype{Y}{>{\centering\arraybackslash}X}

\newcommand{\numRealTasks}{16}
\newcommand{\numScoredRealTasks}{15}

\begin{document}
\title{Pixel Code Policy: Discovering Simple Visuomotor Programs from Demonstrations}

\author{
Hanxiao Jiang$^{1,2}$\textsuperscript{*}\quad
Chenhao Hong$^{1}$\textsuperscript{*}\quad
Qinxi Yu$^{2}$\quad Yuwen Gong$^{1}$\quad
Shenlong Wang$^{2}$\quad Yunzhu Li$^{1}$ \\
\small{$^{1}$Columbia University\quad
$^{2}$University of Illinois Urbana-Champaign} \\
\small{\textsuperscript{*}Equal contribution.} \\
{\small\hypersetup{urlcolor=magenta}\url{https://jianghanxiao.github.io/pixel-code-policy/}}
}

\maketitle
\etocsettocdepth.toc{none} %

\begin{strip}
    \centering
    \vspace{-30pt}
    \includegraphics[width=\linewidth]{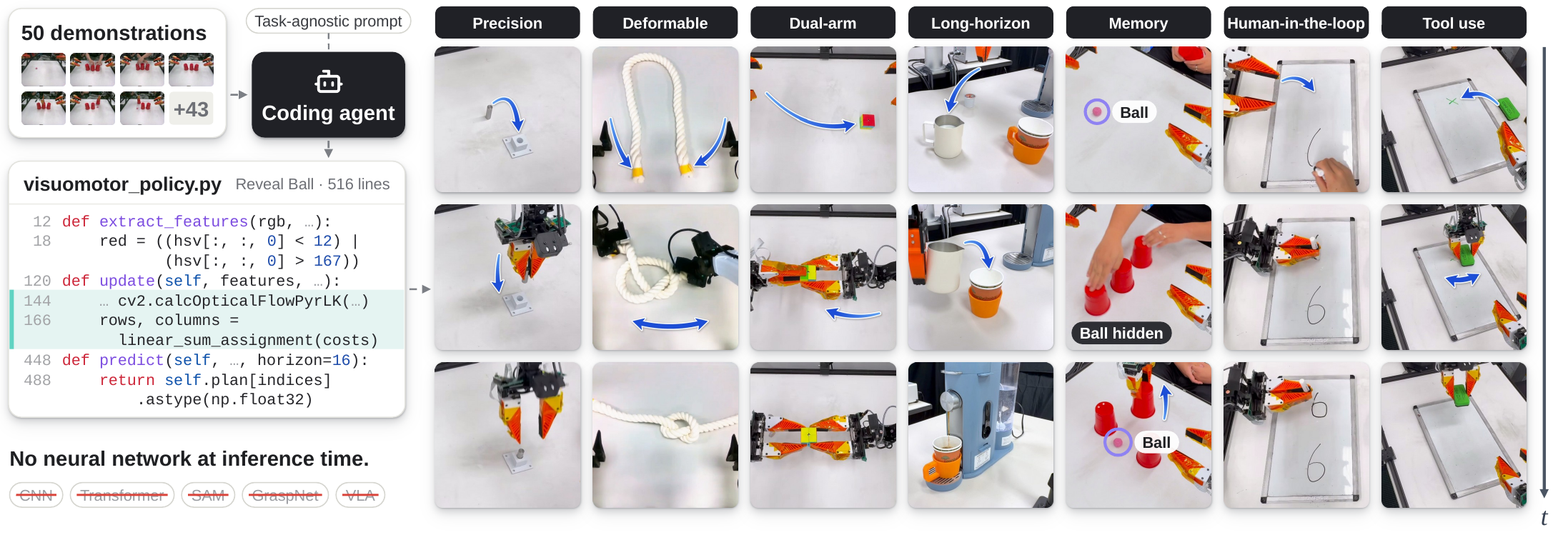}\par
    \par\vspace{-10pt}
\captionof{figure}{\small
    \textbf{Pixel Code Policy solves a diverse range of tasks without any use of neural networks at inference time.}
    \textbf{Left:} From dozens of expert demonstrations, a coding agent discovers task-relevant visual features, fits the corresponding robot motions, and writes them into a pure-code visuomotor policy.
    \textbf{Right:} The policies perform precise, deformable, dual-arm, long-horizon, memory-dependent, human-in-the-loop, and tool-use manipulation on real robots. Each column shows one rollout over time ($t$).
    Each policy call takes less than 19\,ms on a CPU.
}

    \label{fig:teaser}
    \par\vspace{-6pt}
\end{strip}

\begin{abstract}
Learning from demonstrations has enabled robots to perform complex, contact-rich manipulation, largely through neural visuomotor policies. But are neural networks absolutely necessary at inference time? Recent agentic approaches offer alternatives: online agents reason flexibly but incur high inference latency, while code-based approaches often rely on predefined perception and control interfaces. We introduce \textbf{Pixel Code Policy}, an imitation learning framework that uses coding agents to discover surprisingly simple visuomotor programs from dozens of expert demonstrations. The agent autonomously generates programs that identify robust, task-relevant visual features in RGB images, fits the corresponding fine-grained robot motions, and organizes both with task logic and internal state into a pure-code policy. The resulting policies perform precise insertion, deformable-object manipulation, dual-arm coordination, long-horizon execution, and memory-guided interaction, taking less than 19\,ms per call on a CPU, \emph{without any use of neural networks at inference time}. On \numScoredRealTasks{} real-world tasks and all 50 tasks of the RoboTwin 2.0 benchmark, they achieve higher average success than the compared neural policies. Because they are explicit code, the policies can also be analyzed in detail and adapted to new conditions with a single demonstration or a text instruction. Together, these results highlight the potential of agent-discovered programs as a new basis for robot imitation learning, one that is fast, interpretable, and easy to adapt.
\end{abstract}

\IEEEpeerreviewmaketitle

\section{Introduction}
\label{sec:introduction}

Recent advances in visuomotor imitation learning have enabled neural policies to perform complex, contact-rich manipulation~\cite{act,dp,ze20243d,zhang2025flowpolicy}. Vision-language-action and world-action models further extend these capabilities across tasks and environments~\cite{octo,openvla,pi0,pi05,yang2025egovla,kim2026cosmos,ye2026world,yu2026box,motus,UVA}. Relying on neural networks at inference time, however, can come with compromises: large models can be slow to run, and the reasons behind their predicted actions are often hard to interpret. This raises a question: are neural networks necessary at inference time, or can simpler, executable programs learn the same behaviors from demonstrations?

Coding agents offer a way to explore this question by writing executable robot programs, such as those in Code as Policies~\cite{liang2023code} and its successors~\cite{huang2023voxposer,huang2024rekep,fu2026cap}. Unlike online agents, which call models repeatedly during execution~\cite{huang2022inner,chen2026amp,openrua}, these programs run without further agent calls. However, they typically compose predefined perception and motion interfaces, such as segmentation models (e.g., SAM~\cite{kirillov2023sam} and Grounded-SAM~\cite{ren2024groundedsam}), grasp predictors (e.g., GraspNet~\cite{fang2020graspnet}), and motion primitives~\cite{liang2023code,fu2026cap,encore}. These general-purpose perception and planning models are not always reliable for a specific task, and the programs can use only the visual information and motions that the interfaces expose. Dexterous, contact-rich tasks such as knot tying and peg insertion therefore remain challenging. The key challenge is to discover both the task-relevant visual features and the fine-grained robot motions they require, rather than relying on interfaces designed in advance.

We introduce \textbf{Pixel Code Policy}, an imitation learning framework that uses coding agents to discover visuomotor programs from dozens of expert demonstrations (\cref{fig:teaser}). The agent identifies reliable, task-relevant visual features in the demonstrations and fits the corresponding robot motions. Its understanding is fine-grained in both perception and motion. Code-as-policy approaches often reason over object-level abstractions, such as object masks and grasp poses, which can limit fine-grained visual understanding and motion generation; our programs instead measure the image quantities that each motion needs and reproduce the dense motions of the demonstrations. The agent organizes these features and motions with task logic and internal state into a pure-code policy. During execution, the policy uses current observations and robot state to track progress, keep reliable references, and adjust its motions, without any neural network.

We evaluate Pixel Code Policy on \numRealTasks{} real-world tasks across three robot setups, spanning precise insertion, deformable-object manipulation, dual-arm coordination, long-horizon execution, memory, human-in-the-loop interaction, and tool use, and on all 50 tasks of RoboTwin 2.0~\cite{robotwin,robotwin2}. In both settings, the generated programs achieve higher average success than the compared neural policies. Our contributions are threefold:
\begin{enumerate}
    \item For the first time, we show a framework that automatically generates imitation policies that do not rely on any neural networks at inference time, yet reliably perform diverse contact-rich manipulation tasks.
    \item The generated policies support generalization and adaptation: they generalize beyond the demonstrated sequences and adapt to new visual conditions and task requirements with a single demonstration or a text instruction.
    \item By representing each policy as explicit code, our formulation provides the policy with interpretability, supporting detailed analysis and interpretation of what is going on inside each policy, from its visual features and motions to its task logic.
\end{enumerate}

\begin{figure*}[!t]
    \centering
    \includegraphics[width=\linewidth]{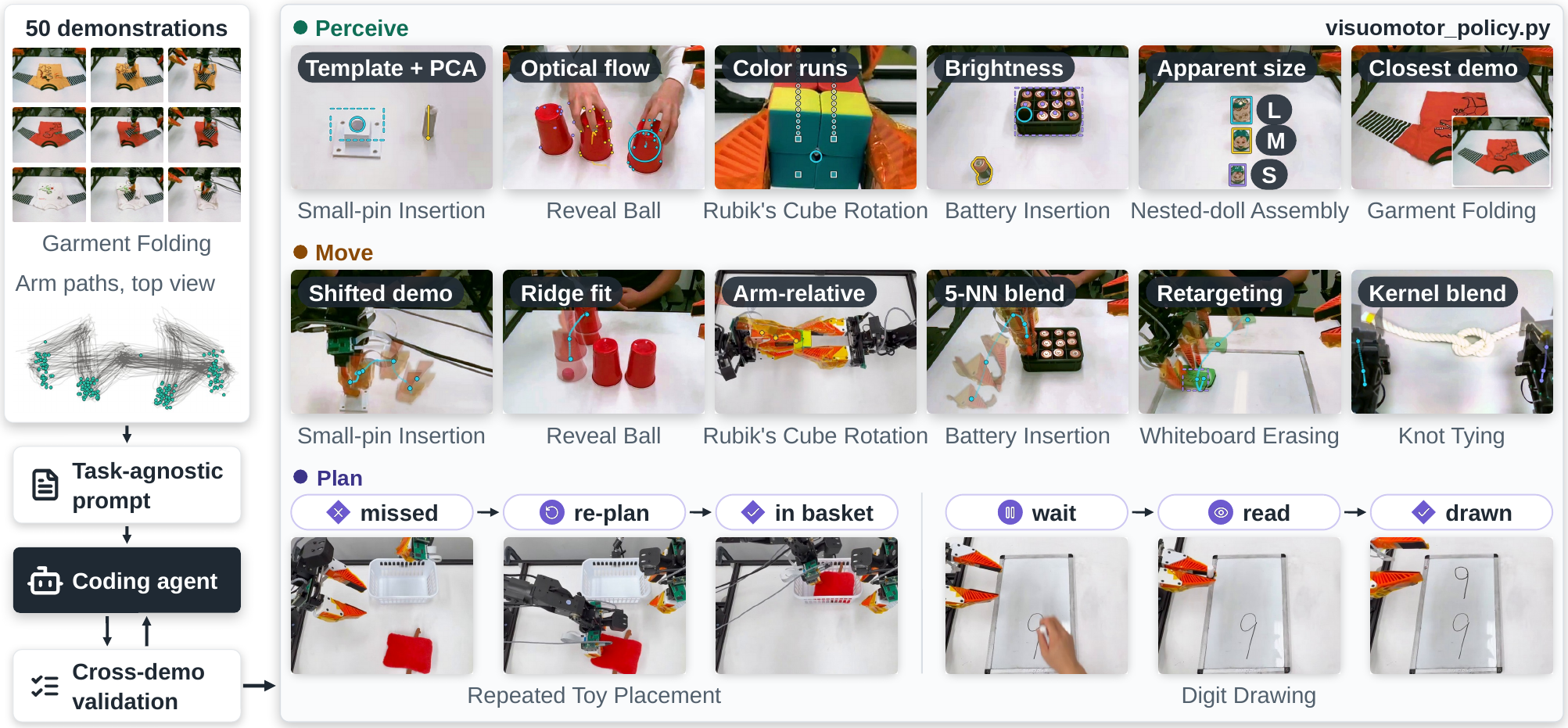}
    \caption{\small
    \textbf{Overview of Pixel Code Policy.}
    Given dozens of demonstrations (left: garment folding, with the end-effector paths of all 50 demonstrations) and a task-agnostic prompt, a coding agent writes a visuomotor policy and refines it through cross-demonstration validation.
    Examples on the right illustrate the three parts: visual feature extraction (\emph{Perceive}), motion generation (\emph{Move}), and task planning (\emph{Plan}).
}
    \label{fig:method}
        \par\vspace{-6pt}
\end{figure*}

\section{Related Work}
\label{sec:related_work}

\textbf{Neural Visuomotor Policies.} 
Task-specific methods such as ACT~\cite{act}, Diffusion Policy~\cite{dp}, 3D Diffusion Policy~\cite{ze20243d}, and flow-based policies~\cite{zhang2025flowpolicy,yan2025maniflow} learn detailed robot motions from demonstrations via neural networks. 
Recent vision-language-action models~\cite{brohan2023rt,octo,openvla,liu2025rdt,pi0,pi05,yang2025egovla} extend this paradigm across diverse tasks and embodiments. 
World-action models~\cite{kim2026cosmos,ye2026world,yu2026box,yuan2026fast,motus,lingbotva,UVA} further learn to jointly predict future states and actions. 
However, these policies remain highly sensitive to camera placement and observation design~\cite{mandlekar2021matters,jiang2026you}. 
While approaches like VINN~\cite{pari2021surprising} demonstrate that strong visual representations can enable effective imitation even through nearest-neighbor retrieval, they still lack explicit logical structure.

\textbf{Agentic Robot Learning.} 
Recent literature explores agentic workflows to synthesize executable robot policies. Inner Monologue~\cite{huang2022inner} repeatedly calls language models during execution to adapt to dynamic environments.
AMP1~\cite{chen2026amp} continuously queries a generalist model to write and execute Python within an episode. 
However, online querying brings high inference latency while many robot tasks do not require heavy reasoning at runtime. 
Alternatively, Code as Policies~\cite{liang2023code} relies heavily on predefined perception APIs, which obscures the agent's raw visual grounding and limits interpretability. CaP-X~\cite{fu2026cap} depends on predefined control APIs and fixed motion abstractions, which inherently abstract away contact-rich movement details and result in low control precision. \textbf{Pixel Code Policy} operates offline, compiling task-relevant visual features and fine-grained action mappings into executable visual-motor logic ahead of time. This allows the policy to run entirely on a CPU without online agent calls or fixed APIs.

\textbf{Agentic Imitation Learning.}
Many recent methods utilize an agentic framework for imitation learning. With zero demonstrations, RAX and ASPIRE~\cite{rax,aspire} synthesize skills through active online exploration, while Act-Observe-Rewrite~\cite{kumar2026actobserverewrite} iteratively modifies controller code across trials. ENPIRE~\cite{enpire} uses coding agents to tune how a policy is trained on real robots. WICL~\cite{roboscript} rehearses tasks in simulation to mitigate physical risks, but this reliance on environment access remains computationally expensive and highly susceptible to the sim-to-real gap. ACE-CaP-0~\cite{ace_cap} develops code policies through trial and error, requesting demonstrations and simulation as needed, then freezes them for execution. RAPID~\cite{rapid} infers task specifications from a single human video. ENCORE~\cite{encore} derives executable policies from few-shot demonstrations. However, relying on such sparse data forces the agent to compensate for task ambiguities through excessive online exploration. 
In contrast, \textbf{Pixel Code Policy} adopts a standard imitation learning setup with dozens of expert demonstrations. Leveraging richer demonstration data enables our coding agent to jointly discover robust visual features and fit dense action mappings for contact-rich manipulation tasks offline.

\section{Method}
\label{sec:method}

Pixel Code Policy gives a coding agent a unified, task-agnostic prompt (\cref{app:prompt}) and the demonstrations of a task. The same prompt is used for all tasks. The agent writes a visuomotor policy as code (\cref{fig:method}). It receives no task-specific perception models or motion primitives.

\subsection{Problem Formulation}
\label{sec:problem}

We follow the standard imitation learning setting. Each task provides $N$ expert demonstrations $\mathcal{D}=\{\tau^i\}_{i=1}^{N}$. Each demonstration $\tau^i=\{(o^i_t,s^i_t,a^i_t)\}_{t=1}^{T_i}$ records RGB images $o_t$, robot states $s_t$ (end-effector poses and gripper openings), and action targets $a_t$. The policy also keeps an internal memory $m_t$. At each call, it maps the current images, state, and memory to the next $H$ action targets and an updated memory:
\begin{equation}
    \big(\hat{a}_{t:t+H},\, m_{t+1}\big)=\pi\big(o_t,s_t,m_t\big).
    \label{eq:policy}
\end{equation}
Neural imitation learning represents $\pi$ by a network $\pi_\theta$. It trains $\pi_\theta$ by minimizing the behavior cloning loss $\mathcal{L}(\pi_\theta;\mathcal{D})$, which sums an action loss $\ell$ over all time steps of the demonstrations (we omit the memory for brevity):
\begin{equation}
    \mathcal{L}(\pi;\mathcal{D})=\sum_{\tau^i\in\mathcal{D}}\,\sum_{t=1}^{T_i}\ell\big(\pi(o^i_t,s^i_t),\,a^i_{t:t+H}\big).
    \label{eq:bc}
\end{equation}
Every call of $\pi_\theta$ then requires network inference. We instead represent the policy as a program $\pi_{c,\phi}$. It consists of code $c$ written by the coding agent and a small set of numerical parameters $\phi$, such as image templates, color thresholds, and motion regression coefficients. The agent fits $\phi$ on one subset of the demonstrations, $\mathcal{D}_{\text{fit}}$, and tests the program on the rest, $\mathcal{D}_{\text{held}}$. It looks for code that also reproduces these held-out demonstrations:
\begin{equation}
    \min_{c}\;\mathcal{L}\big(\pi_{c,\phi_c};\mathcal{D}_{\text{held}}\big),\;\;
    \phi_c=\arg\min_{\phi}\,\mathcal{L}\big(\pi_{c,\phi};\mathcal{D}_{\text{fit}}\big).
    \label{eq:program}
\end{equation}
Searching directly over all possible code is intractable. The agent therefore divides each task into stages and writes the program in three parts:
\begin{equation}
    \begin{gathered}
    z_t=f(o_t,m_t),\qquad (k_t,\,m_{t+1})=g(z_t,s_t,m_t),\\
    \hat{a}_{t:t+H}=h_{k_t}(z_t,s_t;\phi).
    \end{gathered}
    \label{eq:decomposition}
\end{equation}
The perception function $f$ extracts task-relevant visual features $z_t$. The plan function $g$ determines the current stage $k_t$ and updates the memory to $m_{t+1}$. The motion function $h_k$ generates the action chunk for stage $k$. These three parts correspond to \emph{Perceive}, \emph{Plan}, and \emph{Move} in \cref{fig:method}.

\subsection{Generating Visuomotor Programs}
\label{sec:policy_discovery}

Following a lightweight structure inspired by task and motion planning (TAMP)~\cite{tamp}, the agent first splits the task into stages. For each stage, the agent identifies the motion, its geometric constraints, and the expected outcome. Demonstrations differ in speed and duration, so the same stage happens at different frames in different demonstrations. The agent therefore aligns each stage across demonstrations by visual observations, robot motions, and key interaction events such as grasps and releases, not by frame indices. Alignment lets the agent compare the same stage across scene configurations. This comparison shows how the robot motion changes with the visual features, which guides how the agent writes the perception function $f$, the motion functions $h_k$, and the plan function $g$.

\textbf{Perceive.}
The agent does not reconstruct the full scene. It looks only for the visual features that the motions need. It decides what to measure, when to measure it, and how accurately. It builds the perception function $f$ from classical vision techniques~\cite{opencv}, such as color segmentation~\cite{cheng2001color}, template matching~\cite{lewis1995fast}, contour analysis~\cite{suzuki1985topological}, principal component analysis~\cite{pearson1901lines}, optical flow~\cite{lucas1981iterative}, Hough transforms~\cite{duda1972hough}, and feature matching~\cite{lowe2004sift,fischler1981ransac}. These techniques have two advantages. They run in milliseconds on a CPU, and their intermediate results are easy to interpret. The agent can inspect each step, find where an error comes from, and fix or replace that step. This is difficult with black-box neural models. For example, the Small-pin Insertion program finds the hole by template matching and the pin axis by principal component analysis. The Reveal Ball program tracks the cups with optical flow, so it knows which cup hides the ball (\cref{fig:program_analysis} and \cref{fig:program_analysis_supp1}). If an object is briefly occluded during execution, the program keeps its last reliable features. It updates them when a new observation is reliable.

\textbf{Move.}
For each stage, the agent writes a motion function $h_k$ and fits it to the aligned segments of that stage. The agent also chooses which visual features each motion uses and when to measure them. Some motions use features measured once, at the start of the stage. Other motions measure the features again while they run and adjust to reliable new measurements. Classical vision takes only milliseconds, so the program can afford frequent measurements. Each call outputs the next $H=16$ end-effector targets at the demonstration frequency (15\,Hz on our real robots). The program therefore generates dense motions, not only a final pose or a few waypoints. The agent builds each motion function from common fitting operations. Let $\psi$ stack the features and robot state, and let $a^{j}$ and $z^{j}$ be the aligned actions and features of demonstration $j$. \emph{Regression} maps the features to the motion, $\hat{a}_{t:t+H}=W^{\top}\psi(z_t,s_t)$, with $W$ fitted by ridge regression~\cite{hoerl1970ridge}. \emph{Blending} averages the demonstrated motions, $\hat{a}_{t:t+H}=\sum_{j}w_j\,a^{j}_{t:t+H}$, giving larger weights to demonstrations with closer features, $w_j\propto\kappa(\lVert z_t-z^{j}\rVert)$ for a decreasing kernel $\kappa$~\cite{nadaraya1964estimating,watson1964smooth}. \emph{Geometric adaptation} retrieves a demonstration and moves its motion to the current scene, $\hat{a}_{t:t+H}=T\,a^{j}_{t:t+H}$, where the rigid transform $T$ maps the demonstrated object pose to the observed one. Because the program fits dense motions, it captures subtle strategies in the demonstrations. Examples are small contact-rich movements during insertion, choosing an arm based on the object location, and precise coordination between two arms. Each call uses the measured robot state and the current or remembered visual features. The program therefore runs in closed loop and does not just replay a trajectory.

\begin{figure*}[!t]
    \centering
    \includegraphics[width=\linewidth]{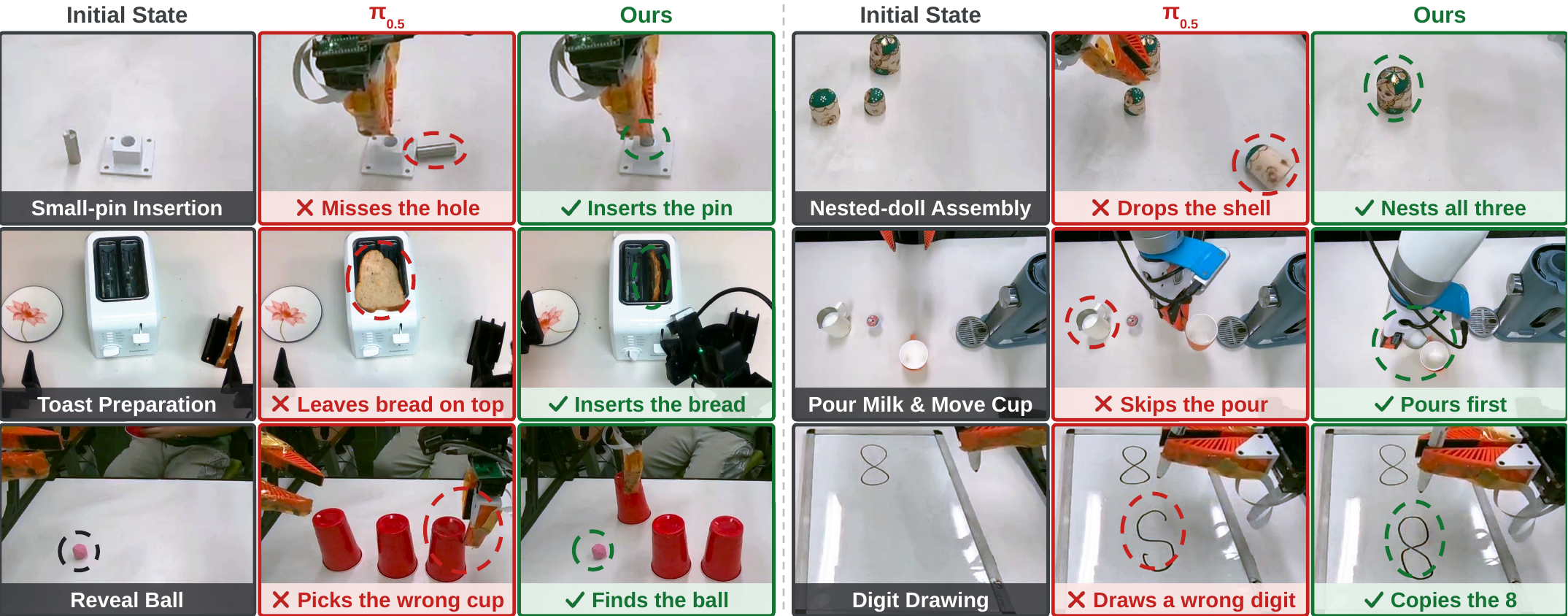}

\caption{\small
    \textbf{Qualitative comparison with $\pi_{0.5}$~\cite{pi05}.}
    For each task, both rollouts start from the initial state on the left. Red dashed ellipses mark where $\pi_{0.5}$ fails, and green dashed ellipses mark where our program succeeds. The black ellipse in the Reveal Ball initial state marks the ball before a cup hides it.
    \textbf{Small-pin Insertion:} insert a narrow pin into a small hole.
    \textbf{Nested-doll Assembly:} fit successive doll shells over smaller pieces.
    \textbf{Toast Preparation:} put bread into a toaster, press the lever, and move the toast to a plate.
    \textbf{Pour Milk \& Move Cup:} pour milk into a cup, then move the cup.
    \textbf{Reveal Ball:} find the cup hiding a ball after a person shuffles the cups.
    \textbf{Digit Drawing:} copy a digit written by a person.
}

    \label{fig:qualitative}
    \par\vspace{-6pt}
\end{figure*}

\textbf{Plan.}
The plan function $g$ follows the same TAMP-inspired structure. From the demonstrations, the agent infers the expected outcome of each stage. The function $g$ checks these outcomes and tracks task progress. It stores information from earlier observations in the memory $m_t$ and selects the next stage. For example, the Nested-doll Assembly program remembers where the dolls are after larger shells cover them (\cref{fig:program_analysis}). The Digit Drawing program remembers the digit a person wrote and decides when to pick up the pen and draw it. The Repeated Toy Placement program detects a failed transfer and retries while the toy is still on the table (\cref{fig:method}). With these outcome checks, the program moves between stages and recovers from failures instead of following a fixed sequence of motions.

\subsection{Cross-Demonstration Validation}
\label{sec:cross_demo_validation}

Cross-demonstration validation is how the agent improves the program toward \cref{eq:program}. Following the prompt, the agent fits the parameters $\phi$ on $\mathcal{D}_{\text{fit}}$. It then runs the program on the held-out demonstrations $\mathcal{D}_{\text{held}}=\mathcal{D}\setminus\mathcal{D}_{\text{fit}}$, using their recorded observations and states. It examines three aspects. For \emph{visual reliability}, the agent checks whether each feature is extracted consistently across demonstrations. A feature that works in one demonstration can fail in others. For example, a color threshold may isolate an object under one lighting condition but not under another. Comparing many demonstrations reveals the most stable features for the task. For \emph{motion accuracy}, the agent measures how far the generated actions are from the demonstrated ones:
\begin{equation}
    e_i=\frac{1}{T_i}\sum_{t=1}^{T_i}\big\lVert\hat{a}^{i}_{t}-a^{i}_{t}\big\rVert_2,\qquad \tau^i\in\mathcal{D}_{\text{held}},
    \label{eq:motion_error}
\end{equation}
where $\hat{a}^{i}_{t}$ is the program's prediction at step $t$ of demonstration $i$. A large $e_i$ may come from unreliable visual features, an unsuitable motion representation, or a wrong task decomposition. For \emph{task logic}, the agent checks whether the plan function recovers the stages and outcomes of each held-out demonstration. The agent analyzes these errors and revises the code. In this way, it improves the task decomposition, perception, motion generation, and planning together. Classical vision and motion fitting are fast and transparent. The agent can therefore test alternatives quickly and trace each error to its source. The process is offline and needs neither the robot nor a simulator. When validation no longer changes the code, the agent refits $\phi$ on all demonstrations, which is a standard practice in cross-validation. The held-out demonstrations therefore shape only the code. All reported results come from separate test trials on the robots, either in the real environment or in simulation.

\subsection{Adapting Programs with Minimal Data}
\label{sec:adaptation_method}

Because the policy is explicit code, the agent can adapt an existing program instead of writing a new one. Besides the program and its original demonstrations, the agent receives either one new demonstration or a short text instruction. It finds the parts of the code that must change and keeps the rest. Often, only the perception code changes. We use this procedure only in the adaptation experiments (\cref{sec:code_adaptation}). There, the agent receives one new demonstration in two cases: a dark room with a moving light (Reveal Ball and Small-pin Insertion) and a towel of a new color (Towel Folding). For Digit Drawing, it receives no new demonstration, only a text instruction to copy handwritten letters instead of digits. The failure-feedback experiments (\cref{sec:exp_results}) use the same procedure with a text description or a video of a failed execution. All other results use programs generated from the original demonstrations alone.

\section{Experiments}
\label{sec:experiments}

We evaluate Pixel Code Policy on \numRealTasks{} real-world manipulation tasks across three robot setups and all 50 tasks in RoboTwin 2.0. Our primary objective is to address the following four questions: (1)~How well do the generated programs perform compared with neural visuomotor policies? (2)~What do the programs learn from demonstrations, how does the number of demonstrations affect their performance, and can they generalize beyond the demonstrated sequences? (3)~Can failure feedback improve the generated programs on challenging tasks? (4)~Can the generated programs be adapted to new visual conditions and task requirements through targeted code changes?

\subsection{Experimental Settings}
\label{sec:exp_settings}

\textbf{Implementation Details.}
We use GPT-6.1 Sol~\cite{gpt61sol} in Ultra mode as the coding agent, except for Knot Tying, Toast Preparation, and the full Coffee Preparation sequence, which use GPT-6.2 Sol~\cite{gpt62sol}. All tasks share the same task-agnostic prompt, provided in \cref{app:prompt}. For each task, the coding agent independently generates three candidate programs in parallel using only the recorded demonstrations, without interacting with the robot or simulator. We select the best candidate based on validation trials and evaluate it on separate test trials without further code changes. Across the 13 ALOHA tasks, generating the selected program takes an average of 3.8 hours and 112 million tokens, while fine-tuning $\pi_{0.5}$ takes an average of 7.4 hours on a single NVIDIA RTX PRO 6000 Blackwell GPU with 96\,GB of memory. During deployment, each policy call predicts a chunk of 16 action targets, which are executed at 15\,Hz on the real robots, and takes less than 19\,ms on a CPU (AMD Ryzen 9 7950X), excluding low-level robot control and motion planning.

\begin{table}[!t]
    \caption{\small \textbf{Real-world tasks and their associated capabilities.}
    \textbf{P}: precision;
    \textbf{D}: deformable objects;
    \textbf{B}: dual-arm;
    \textbf{L}: long-horizon;
    \textbf{M}: memory;
    \textbf{H}: human-in-the-loop;
    \textbf{T}: tool use.}
    \label{tab:real_task_matrix}
    \centering
    \begingroup
    \setlength{\tabcolsep}{3.2pt}
    \renewcommand{\arraystretch}{0.98}
    \footnotesize
    \newcommand{\tmyes}{\checkmark}
    \begin{tabular*}{\columnwidth}{@{\extracolsep{\fill}}l*{7}{c}@{}}
        \toprule
        Task & P & D & B & L & M & H & T \\
        \midrule
        Small-pin Insertion    & \tmyes &        &        &        &        &        &        \\
        Battery Insertion      & \tmyes &        &        &        &        &        &        \\
        Mahjong Matching       & \tmyes &        &        &        &        &        &        \\
        Nested-doll Assembly   & \tmyes &        & \tmyes &        & \tmyes &        &        \\
        \addlinespace[2pt]
        Towel Folding          &        & \tmyes & \tmyes &        &        &        &        \\
        Garment Folding        &        & \tmyes & \tmyes &        &        &        &        \\
        Rubik's Cube Rotation  & \tmyes &        & \tmyes &        &        &        &        \\
        Toy Packing            &        & \tmyes & \tmyes &        &        &        &        \\
        Knot Tying             &        & \tmyes & \tmyes & \tmyes &        &        &        \\
        \addlinespace[2pt]
        Screwdriver Handover   &        &        & \tmyes &        &        &        &        \\
        Whiteboard Erasing     &        &        & \tmyes &        &        &        & \tmyes \\
        Toast Preparation      &        & \tmyes & \tmyes & \tmyes &        &        &        \\
        Coffee Preparation     &        &        &        & \tmyes &        &        &        \\
        \addlinespace[2pt]
        Reveal Ball            &        &        &        &        & \tmyes & \tmyes &        \\
        Digit Drawing          &        &        &        &        & \tmyes & \tmyes & \tmyes \\
        Repeated Toy Placement &        & \tmyes &        & \tmyes & \tmyes & \tmyes &        \\
        \bottomrule
    \end{tabular*}
    \par\vspace{-10pt}
    \endgroup
\end{table}

\begin{table*}[!t]
    \caption{\small
    \textbf{Real-world manipulation success rates.}
    Both methods use the same 20 initial object placements per task.
    }
    \label{tab:real_results}
    \centering
    \begingroup
    \footnotesize
    \setlength{\tabcolsep}{2.5pt}
    \renewcommand{\arraystretch}{1.18}
    \begin{tabularx}{\textwidth}{@{}>{\raggedright\arraybackslash}m{0.19\textwidth}*{8}{Y}@{}}
        \toprule
        Method & \makecell{Small-pin\\Insertion} & \makecell{Battery\\Insertion} & \makecell{Mahjong\\Matching} & \makecell{Nested-doll\\Assembly} & \makecell{Towel\\Folding} & \makecell{Garment\\Folding} & \makecell{Rubik's Cube\\Rotation} & \makecell{Toy\\Packing} \\
        \midrule
        $\pi_{0.5}$ & 5/20 & 13/20 & 1/20 & 1/20 & 15/20 & 16/20 & 4/20 & 8/20 \\
        \textbf{Pixel Code Policy (Ours)} & \textbf{19/20} & \textbf{17/20} & \textbf{15/20} & \textbf{17/20} & \textbf{20/20} & \textbf{19/20} & \textbf{13/20} & \textbf{14/20} \\
        \bottomrule
    \end{tabularx}
    \par\vspace{5pt}
    \begin{tabularx}{\textwidth}{@{}>{\raggedright\arraybackslash}m{0.19\textwidth}*{8}{Y}@{}}
        \toprule
        Method & \makecell{Screwdriver\\Handover} & \makecell{Whiteboard\\Erasing} & \makecell{Toast\\Preparation} & \makecell{Pour Milk\\\& Move Cup} & \makecell{Reveal\\Ball} & \makecell{Digit\\Drawing} & \makecell{Repeated Toy\\Placement} & \makecell{Mean\\(\%)} \\
        \midrule
        $\pi_{0.5}$ & 8/20 & 2/20 & 14/20 & 12/20 & 2/20 & 9/20 & 15/20 & 41.7 \\
        \textbf{Pixel Code Policy (Ours)} & \textbf{18/20} & \textbf{14/20} & \textbf{19/20} & \textbf{13/20} & \textbf{17/20} & \textbf{20/20} & \textbf{19/20} & \textbf{84.7} \\
        \bottomrule
    \end{tabularx}
     \par\vspace{-6pt}
    \endgroup
\end{table*}

\begin{table}[!t]
    \caption{\small
    \textbf{RoboTwin 2.0 Clean-to-Clean results.}
    100 test scenes per task.
    $^{\dagger}$ Uses depth; $^{\ddagger}$ co-trained on all 50 tasks.
    }
    \label{tab:sim_results}
    \centering
    \begingroup
    \setlength{\tabcolsep}{2pt}
    \renewcommand{\arraystretch}{1.08}
    \footnotesize
    \begin{tabular*}{\columnwidth}{@{\extracolsep{\fill}}lrrrrr@{}}
        \toprule
        Task & DP & ACT & DP3$^{\dagger}$ & $\pi_{0.5}^{\ddagger}$ & Ours \\
        \midrule
        Blocks Ranking RGB & 0\% & 1\% & 3\% & 72\% & \textbf{100\%} \\
        Place Burger Fries & 72\% & 49\% & 72\% & 82\% & \textbf{100\%} \\
        Move Playingcard Away & 47\% & 36\% & 68\% & 84\% & \textbf{99\%} \\
        Lift Pot & 39\% & 88\% & 97\% & \textbf{98\%} & \textbf{98\%} \\
        Place Cans Plasticbox & 40\% & 16\% & 48\% & 31\% & \textbf{95\%} \\
        Handover Block & 10\% & 42\% & 70\% & 30\% & \textbf{93\%} \\
        Open Laptop & 49\% & 56\% & 82\% & \textbf{93\%} & 81\% \\
        Move Stapler Pad & 1\% & 0\% & 12\% & 20\% & \textbf{72\%} \\
        Place A2B Left & 2\% & 1\% & 46\% & \textbf{71\%} & 69\% \\
        \midrule
        Mean (50 tasks) & 28.04\% & 29.74\% & 55.24\% & 70.66\% & \textbf{78.24\%} \\
        \bottomrule
    \end{tabular*}
     \par\vspace{-10pt}
    \endgroup
\end{table}

\textbf{Real-World Tasks.}
We evaluate \numRealTasks{} real-world manipulation tasks covering seven capabilities: precision manipulation, deformable-object manipulation, dual-arm coordination, long-horizon execution, memory, human-in-the-loop interaction, and tool use (\cref{tab:real_task_matrix}). Coffee Preparation is performed on a single-arm xArm, Toast Preparation and Knot Tying on a dual-arm YAM, and the remaining 13 tasks on a dual-arm ALOHA. For each task, we collect 50 expert demonstrations at 15\,Hz, recording RGB observations from a head camera and a camera on each wrist, together with robot states and action targets. All camera views are available to the coding agent, which autonomously determines which views are useful for the task. We quantitatively evaluate \numScoredRealTasks{} tasks over 20 trials per task, using the same initial object placements for both methods. We further explore learning from failure feedback on Knot Tying and Pour Milk. Detailed task descriptions and robot setups are provided in \cref{app:real_task_coverage}.

\textbf{Real-World Baseline.}
We compare Pixel Code Policy with $\pi_{0.5}$~\cite{pi05}, the strongest neural policy among our RoboTwin baselines. We fully fine-tune $\pi_{0.5}$ from the official \texttt{pi05\_base} checkpoint for 10{,}000 steps with a batch size of 32 in BF16 on a single 96\,GB RTX PRO 6000 Blackwell GPU. The policy receives RGB observations from all available cameras and the end-effector state, and predicts 30-step action chunks. We adopt training-time real-time chunking (RTC) with delays of 0--8 steps~\cite{black2025rtc,black2025trainingrtc}.

\textbf{RoboTwin Benchmark.}
We evaluate all 50 tasks in RoboTwin 2.0~\cite{robotwin2} following the standard Clean-to-Clean setting, using 50 expert demonstrations and the same 100 test seeds as the official benchmark. We select the best of three generated programs using five separate validation seeds that do not overlap with the test seeds. We compare with DP, ACT, DP3, and the co-trained $\pi_{0.5}$ using the official leaderboard results.

\subsection{Results}
\label{sec:exp_results}

\textbf{Comparison with Neural Policies.}
On the real robots, Pixel Code Policy achieves an average success rate of 84.7\% across the \numScoredRealTasks{} quantitatively evaluated tasks, compared with 41.7\% for $\pi_{0.5}$(\cref{tab:real_results}). With only 50 expert demonstrations per task, the policies must handle substantial variations in object positions, orientations, and scene configurations. Many tasks also require precise, contact-rich motions or memory of information from earlier observations, making them particularly challenging for neural policies to learn from limited demonstrations. The performance gap is especially pronounced in precision manipulation, where our programs achieve 81\% success compared with 24\% for $\pi_{0.5}$, and in tasks requiring memory, where the corresponding success rates are 91\% and 34\%.

\Cref{fig:qualitative} shows representative rollouts starting from the same initial object placements. In Small-pin Insertion and Nested-doll Assembly, $\pi_{0.5}$ struggles with precise physical interactions, while our programs use task-relevant geometric features to generate accurate motions. In Toast Preparation and Pour Milk \& Move Cup, $\pi_{0.5}$ misses essential steps, whereas our programs maintain task progress through explicit task logic. In Reveal Ball and Digit Drawing, the programs retain information from earlier observations and use it to guide subsequent actions, even when the relevant information is no longer visible.

On RoboTwin, Pixel Code Policy outperforms all compared neural policies in overall success rate (\cref{tab:sim_results}). Our programs outperform the co-trained $\pi_{0.5}$ on 28 of the 50 tasks and tie on four, without depth observations or neural network inference. Complete per-task results are provided in \cref{app:sim_results}.

\begin{figure*}[!t]
    \centering
    \includegraphics[width=\linewidth]{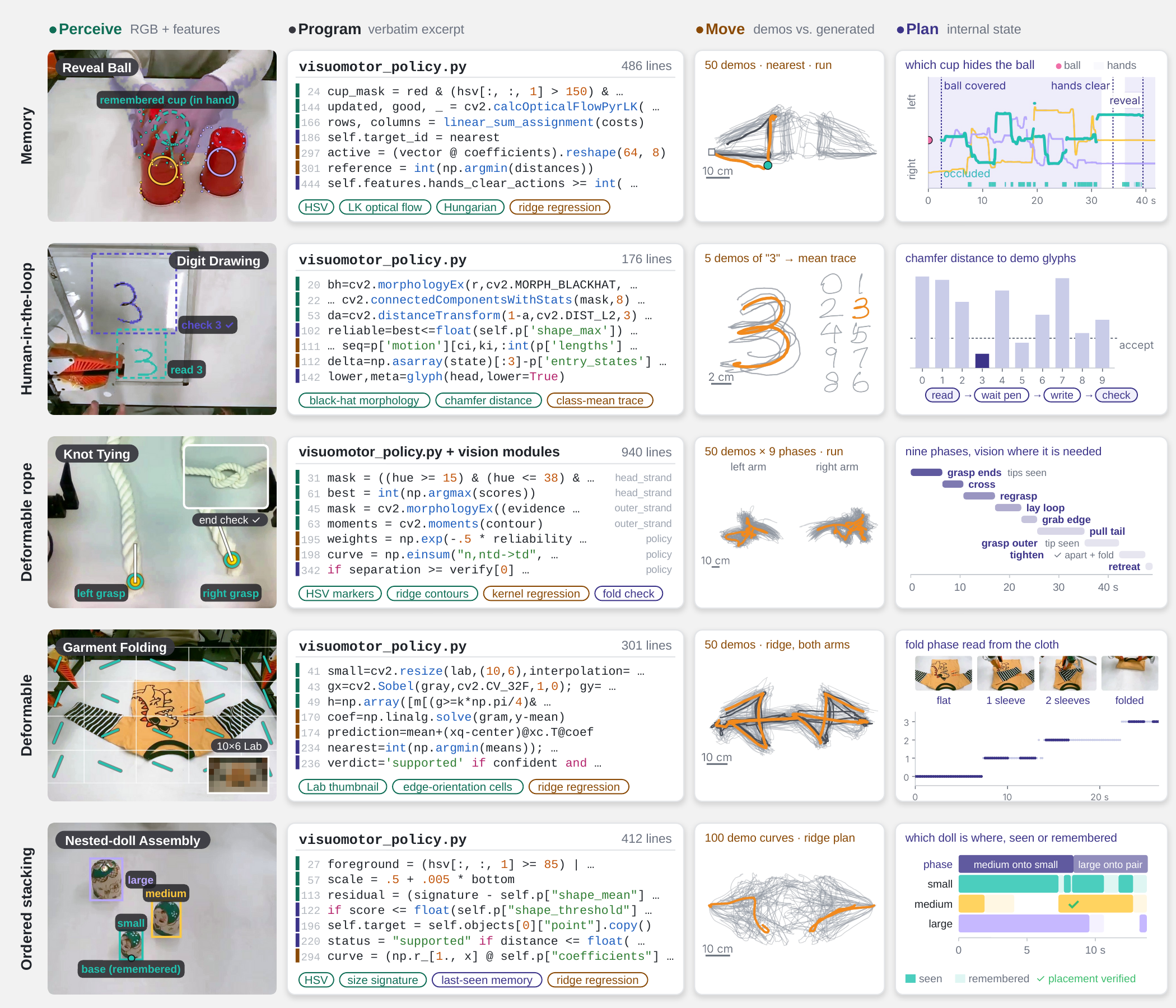}
    \caption{\small
    \textbf{Analysis of generated visuomotor programs.}
    Each row shows a real-world task and the capability it highlights.
    \textbf{Perceive:} Head-camera images overlaid with extracted visual features.
    \textbf{Program:} Code excerpts for perception, motion generation, and planning.
    \textbf{Move:} Demonstrated (gray), retrieved (dark gray), and generated (orange) end-effector trajectories.
    \textbf{Plan:} Task progress and internal state.
}
    \label{fig:program_analysis}
    \par\vspace{-10pt}
\end{figure*}

\textbf{Program Analysis.}
To understand how the generated programs achieve these behaviors, we visualize their discovered visual features, source code, fitted motions, and internal state in \cref{fig:program_analysis}. In Reveal Ball, the program combines optical-flow tracking with correspondence matching to maintain the identity of the cup hiding the ball. In Nested-doll Assembly, it identifies individual dolls through their size signatures and remembers their locations when they become covered. In Knot Tying, the program organizes coordinated bimanual motions into nine stages and extracts visual features only when they are needed. In Garment Folding, visual observations determine the current folding stage and its associated motions. These examples illustrate how the coding agent jointly discovers task-relevant perception, motion mappings, and task logic instead of relying on predefined task-specific abstractions.

The generated programs are also compact. Across the 50 RoboTwin tasks, their runtime code contains a median of 320 lines, excluding blank lines and comments. Despite their simplicity, the programs employ diverse combinations of classical vision techniques, numerical motion fitting, and task logic. Additional analyses of the remaining real-world programs are provided in \cref{fig:program_analysis_supp1,fig:program_analysis_supp2}.

\begin{figure*}[!t]
    \centering
    \includegraphics[width=\linewidth]{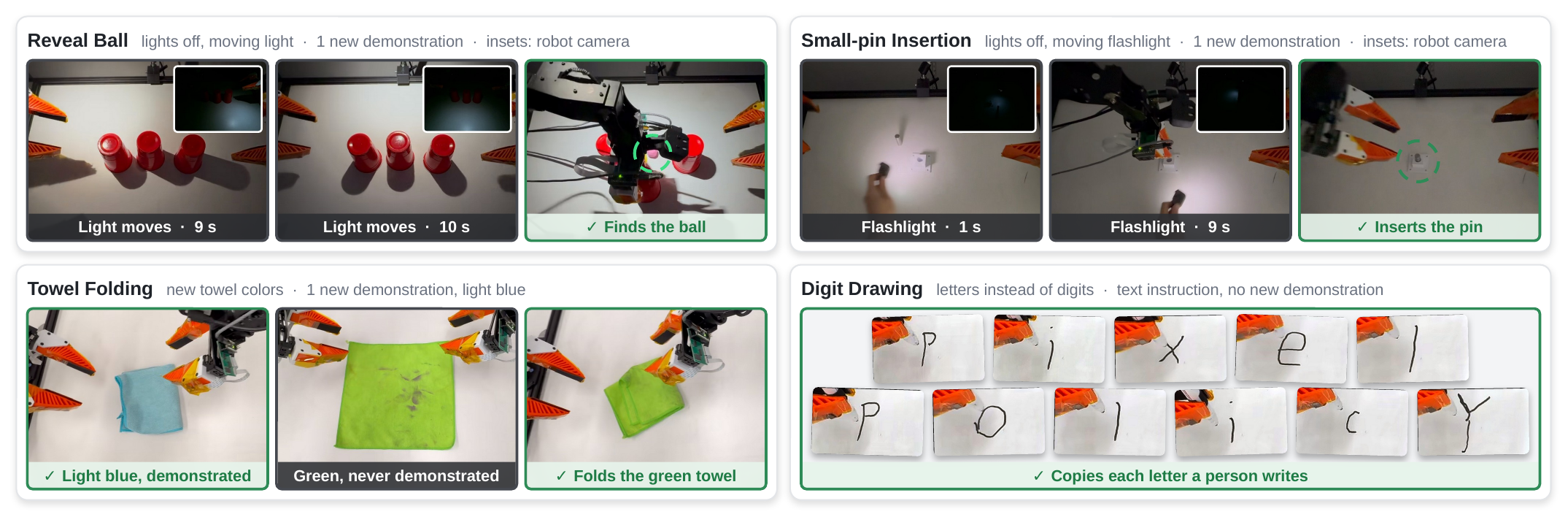}
    \caption{\small
    \textbf{Adapting programs to new conditions.}
    \textbf{Adversarial lighting:} With one additional demonstration per task, the Reveal Ball and Small-pin Insertion programs adapt to a dark room with a moving light source.
    \textbf{Object appearance variations:} Given one demonstration with a light-blue towel, the adapted program folds both light-blue and previously unseen green towels.
    \textbf{New task instructions:} Given only a text instruction, the Digit Drawing program is adapted from copying digits to copying handwritten letters.
}
    \label{fig:adaptation}
    \par\vspace{-6pt}
\end{figure*}

\textbf{Number of Demonstrations.}
Multiple demonstrations are important for both discovering robust visual features and learning accurate, contact-rich motions. We investigate this on Small-pin Insertion using 1, 10, and 50 demonstrations, achieving success rates of 0\%, 50\%, and 95\%, respectively, on the same 20 test placements. With a single demonstration, the agent cannot validate visual features across different scene configurations, and the resulting program fails to extract reliable visual features when the object placement changes. With 10 demonstrations, the agent discovers more robust visual features but still struggles to fit accurate motion mappings due to the limited diversity of demonstrated motions. The generated trajectories may deviate from the subtle, contact-rich movements needed for successful insertion. With 50 demonstrations, the agent can both discover reliable visual features and fit more accurate motions by leveraging a richer set of demonstrated manipulation strategies. These results highlight the importance of multiple demonstrations for jointly improving visual feature extraction and fine-grained motion generation.

\textbf{Generalization Beyond Demonstrations.}
The generated programs exhibit behaviors beyond the demonstrated sequences without any further code changes. In Reveal Ball, although the demonstrations contain only one to three cup exchanges, the program tracks the target cup through much longer shuffling sequences by maintaining its identity and updating its location. In Repeated Toy Placement, the demonstrations contain only one to three placement cycles. The program continuously responds to human interactions, places the toy back whenever it is removed, and retries unsuccessful grasps, sustaining the interaction for ten minutes in one experiment. These examples show how the task logic and internal state discovered from demonstrations allow the programs to generalize beyond the demonstrated temporal horizons rather than simply reproducing observed action sequences.

\textbf{Learning from Failure Feedback.}
All programs evaluated above are generated from expert demonstrations and deployed without further code modifications. We further investigate whether failure feedback can improve policy performance on two challenging tasks: Pour Milk \& Move Cup and Knot Tying. We provide the coding agent with either a text prompt describing the failure or a video of a failed execution. By examining the extracted visual features, generated motions, and task logic, the agent can diagnose failures and revise the corresponding code. With limited failure feedback, the success rate of Pour Milk \& Move Cup increases from 65\% to 80\%, while Knot Tying improves from 0\% to 55\%. These results show that explicit visuomotor programs can be effectively improved through targeted feedback, particularly for challenging tasks involving complex physical interactions and large variations in object states. Our approach is therefore complementary to feedback-driven self-improvement: learning fine-grained, contact-rich motions from expert demonstrations provides a strong foundation for subsequent program refinement, rather than requiring the agent to discover these motions through trial and error.

\subsection{Code Adaptation}
\label{sec:code_adaptation}

Following \cref{sec:adaptation_method}, we adapt existing programs to three kinds of change (\cref{fig:adaptation}). \textbf{Adversarial lighting.} In Reveal Ball and Small-pin Insertion, the robots operate in a dark room with a moving light source. Given one new demonstration, the agent revises the visual feature extraction to account for changing brightness and shadows while largely preserving the existing motion generation and task logic. \textbf{Object appearance.} The original Towel Folding program is generated from demonstrations with a dark-blue towel. Given one demonstration with a light-blue towel, the agent replaces the color-based feature extraction with a background-based approach and adapts the existing folding motions to the extracted cloth geometry. The resulting program folds both the light-blue towel and a previously unseen green towel. \textbf{New task instruction.} Without additional demonstrations, we ask the agent to extend the Digit Drawing program from copying digits to copying handwritten letters. The agent modifies the perception code to reconstruct handwritten strokes by tracking the marker tip and uses the original digit demonstrations to learn when the marker touches the board. The adapted program reproduces all 11 letters of ``Pixel Policy''. In all three cases, the explicit code allows the agent to reuse previously learned behaviors and change only the perception, motion, or task logic affected by the new requirement.

\section{Conclusion}
\label{sec:conclusion}

We introduced Pixel Code Policy, an imitation learning framework that uses coding agents to discover simple visuomotor programs from dozens of expert demonstrations. By discovering task-relevant visual features, fitting the corresponding robot motions, and designing task logic, the agent autonomously generates closed-loop policies that perform diverse, contact-rich manipulation tasks \emph{without neural networks at inference time} or online agent calls. Our programs substantially outperform $\pi_{0.5}$ across \numScoredRealTasks{} real-world tasks and achieve higher average success than all compared neural policies on RoboTwin 2.0. Beyond their performance and computational efficiency, the explicit programs generalize beyond demonstrated sequences and can be improved or adapted with minimal additional data. More broadly, our framework provides a new way of interfacing between agents and robot policies that work reliably and dexterously in the real world. Looking forward, combining demonstration-based program generation with feedback-driven self-improvement offers a promising direction for further improving policy performance. This work suggests that very simple but task-specific visual features, generated automatically and without any use of neural networks, can already be surprisingly effective for complicated, contact-rich tasks.

\section*{Acknowledgments}
This work was partially supported by the Toyota Research Institute, Analog Devices, and NSF Award \#2409661. This project is also supported by NSF Awards \#2331878, \#2340254, \#2312102, \#2414227, and \#2404385. We greatly appreciate the NCSA for providing computing resources. This article solely reflects the opinions and conclusions of its authors and should not be interpreted as necessarily representing the official policies, either expressed or implied, of the sponsors. 

\bibliographystyle{plainnat}
\bibliography{references}

\clearpage
\onecolumn
\raggedbottom
\setcounter{section}{0}
\renewcommand{\thesection}{\Alph{section}}
\renewcommand{\thesectiondis}{\thesection.}
\renewcommand{\thesubsection}{\thesection.\arabic{subsection}}
\renewcommand{\thesubsectiondis}{\thesubsection.}
\renewcommand{\theHsection}{supp.\arabic{section}}
\renewcommand{\theHsubsection}{supp.\arabic{section}.\arabic{subsection}}

\setcounter{figure}{0}
\renewcommand{\thefigure}{S\arabic{figure}}
\renewcommand{\theHfigure}{supp.\arabic{figure}}
\setcounter{table}{0}
\renewcommand{\thetable}{S\arabic{table}}
\renewcommand{\theHtable}{supp.\arabic{table}}
\setcounter{equation}{0}
\renewcommand{\theequation}{S\arabic{equation}}
\renewcommand{\theHequation}{supp.\arabic{equation}}

\section*{Supplementary Material}

\etocsettocdepth.toc{subsection}
\begingroup
\etocsettocstyle{\noindent\textbf{Contents}\par\vspace{-\smallskipamount}\begin{multicols}{2}}{\end{multicols}}
\etocsetnexttocdepth{subsection}
\tableofcontents
\endgroup

\section{Real-World Tasks}
\label{app:real_task_coverage}

\Cref{tab:real_task_capabilities} describes the real-world tasks, grouped as in \cref{tab:real_task_matrix}.

\begin{table}[H]
    \caption{\small \textbf{Real-world task descriptions.}}
    \label{tab:real_task_capabilities}
    \centering
    \begingroup
    \setlength{\tabcolsep}{4pt}
    \renewcommand{\arraystretch}{1.0}
    \footnotesize
    \begin{tabular*}{\textwidth}{@{\extracolsep{\fill}}>{\raggedright\arraybackslash}p{0.14\textwidth}>{\raggedright\arraybackslash}p{0.37\textwidth}>{\raggedright\arraybackslash}p{0.45\textwidth}@{}}
        \toprule
        Task & Operation & Capability evaluated \\
        \midrule
        \multicolumn{3}{@{}l}{\textbf{Precision manipulation and visual target selection}} \\
        \addlinespace[2pt]
        Small-pin Insertion & Insert a narrow metal pin into a small opening. & Precise alignment, contact-rich insertion, and arm choice as the pin moves. \\
        Battery Insertion & Insert a battery into the empty slot of a holder. & Detecting the unoccupied slot and adjusting to its changing location. \\
        Mahjong Matching & Find the loose tile that matches a set of three tiles and stack it on the set. & Comparing tile faces among tiles of the same shape and placing a small tile precisely on top of the set. \\
        Nested-doll Assembly & Fit successive doll shells over smaller pieces. & Repeated assembly and alignment with changing object positions, remembering pieces once they are covered. \\
        \midrule
        \multicolumn{3}{@{}l}{\textbf{Deformable-object and bimanual manipulation}} \\
        \addlinespace[2pt]
        Towel Folding & Fold a towel twice using both arms. & Bimanual folding across varied positions, orientations, and cloth shapes. \\
        Garment Folding & Fold the sleeves and body of a long-sleeved garment. & Coordinated folding across garments with different colors and patterns. \\
        Rubik's Cube Rotation & Hold a small cube with two grippers and turn one layer. & Fine bimanual alignment and relative rotation of the grippers. \\
        Toy Packing & Place a soft toy in a hinged box and close the lid. & Soft-object placement and coordinated manipulation of the box lid. \\
        Knot Tying & Form a knot using coordinated arm motions. & Coordinating two arms over nine phases while manipulating a flexible cord. \\
        \midrule
        \multicolumn{3}{@{}l}{\textbf{Tool use, handover, and long-horizon operation}} \\
        \addlinespace[2pt]
        Screwdriver Handover & Transfer a screwdriver between arms, then place it in a cup. & Mid-air handover and arm selection under changing object placement. \\
        Whiteboard Erasing & Erase marks using a board eraser, with handover as needed. & Tool use and choosing actions from varying mark positions and counts. \\
        Toast Preparation & Put bread into a toaster, press the lever, and move it to a plate. & Handling soft bread, sequential appliance operation, and coordination between arms. \\
        Coffee Preparation & Pour milk and move the cup as part of a multi-stage coffee preparation sequence. & Visually guided pouring and sequential manipulation. \\
        \midrule
        \multicolumn{3}{@{}l}{\textbf{Interaction and temporal reasoning}} \\
        \addlinespace[2pt]
        Reveal Ball & Select the cup hiding a ball after a person shuffles three cups. & Tracking an occluded target using information from earlier observations. \\
        Digit Drawing & Respond to a digit written by a person. & Matching handwritten digit shapes, remembering the digit while the board is occluded, and selecting a visually conditioned response. \\
        Repeated Toy Placement & Return a soft toy to a container each time a person removes it. & Sustaining human--robot interaction, recognizing changed scenes, and retrying failed grasps. \\
        \bottomrule
    \end{tabular*}

    \endgroup
\end{table}

\subsection{Real-Robot Setups}
\label{app:robot_setups}

\Cref{tab:robot_setups} summarizes the three robot setups. Nine of the 13 ALOHA programs use only the head camera; Whiteboard Erasing and Repeated Toy Placement use all three cameras, while Rubik's Cube Rotation and Digit Drawing each use the head camera and one wrist camera.

\begin{table}[H]
    \caption{\small \textbf{Real-robot setups.} All cameras are Intel RealSense.}
    \label{tab:robot_setups}
    \centering
    \begingroup
    \setlength{\tabcolsep}{3pt}
    \renewcommand{\arraystretch}{1.12}
    \footnotesize
    \begin{tabular*}{\textwidth}{@{\extracolsep{\fill}}>{\raggedright\arraybackslash}p{0.07\textwidth}>{\raggedright\arraybackslash}p{0.16\textwidth}>{\raggedright\arraybackslash}p{0.15\textwidth}>{\raggedright\arraybackslash}p{0.24\textwidth}>{\raggedright\arraybackslash}p{0.12\textwidth}>{\raggedright\arraybackslash}p{0.11\textwidth}@{}}
        \toprule
        Setup & Tasks & Robot & Cameras and policy images & Action chunk & Low-level control \\
        \midrule
        xArm & Coffee Preparation & xArm7, single-arm & D455 head camera, D435i wrist camera; $320\times180$ RGB & 16 steps $\times$ 8 values & Joint servo at 50\,Hz \\
        YAM & Toast Preparation, Knot Tying & I2RT YAM, dual-arm & D435 head camera, D405 on each wrist; $320\times240$ RGB & 16 steps $\times$ 16 values & Follower control at 100\,Hz \\
        ALOHA & Other 13 tasks & ALOHA-style~\cite{act}, dual-arm & D455 head camera, D405 on each wrist; $320\times240$ RGB & 16 steps $\times$ 16 values & Inverse kinematics, targets at 15\,Hz \\
        \bottomrule
    \end{tabular*}
    \endgroup
\end{table}

\clearpage
\subsection{Program Analysis Across Real-World Tasks}
\label{app:real_program_analysis}

\begin{figure}[H]
    \centering
    \includegraphics[width=\linewidth,height=0.86\textheight,keepaspectratio]{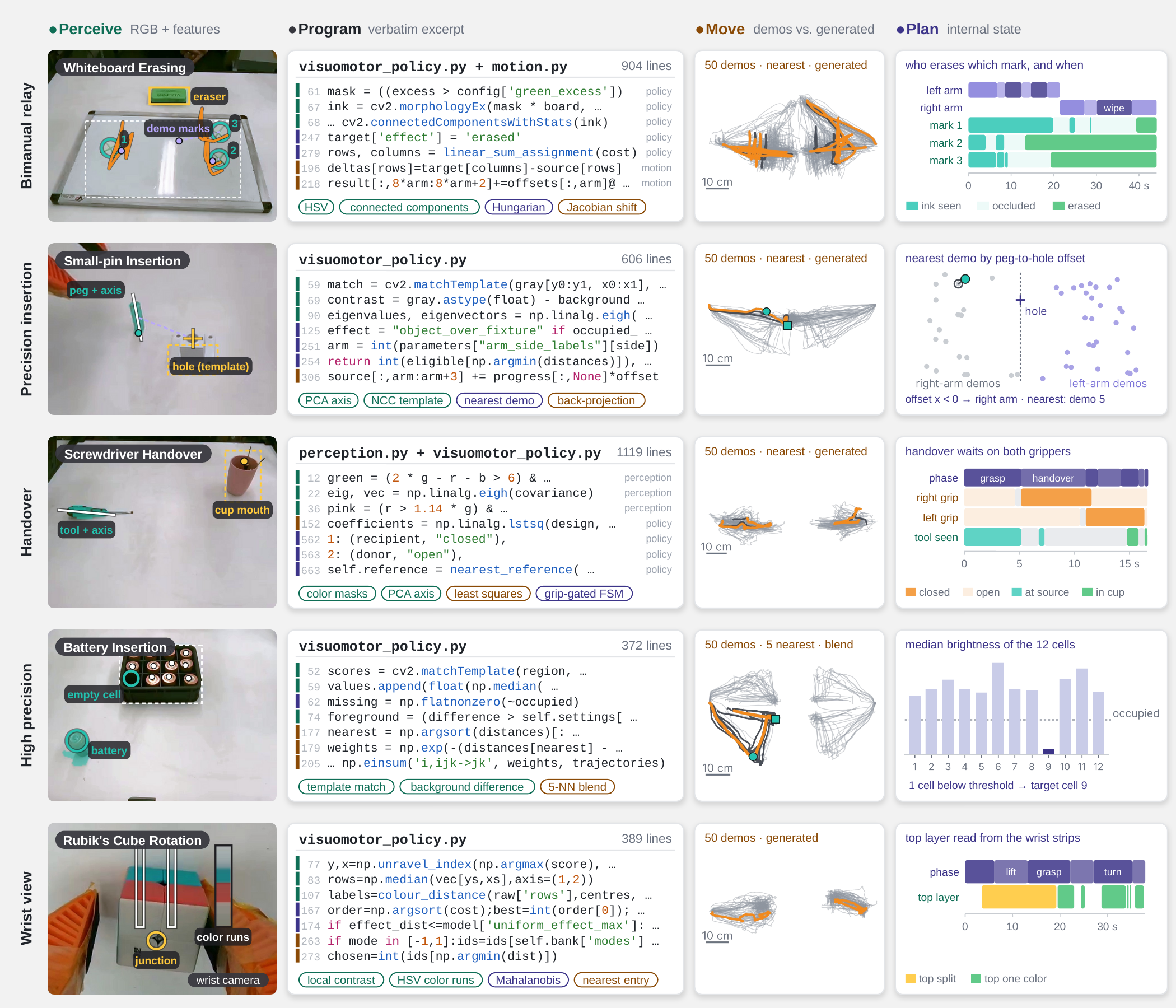}
    \caption{\small \textbf{Additional real-world program analyses (1/2).} Rubik's Cube Rotation uses a wrist-camera view. Generated trajectories and internal-state traces are obtained by replaying the programs on recorded observations under ideal tracking.}
    \label{fig:program_analysis_supp1}
\end{figure}

\begin{figure}[H]
    \centering
    \includegraphics[width=\linewidth,height=0.86\textheight,keepaspectratio]{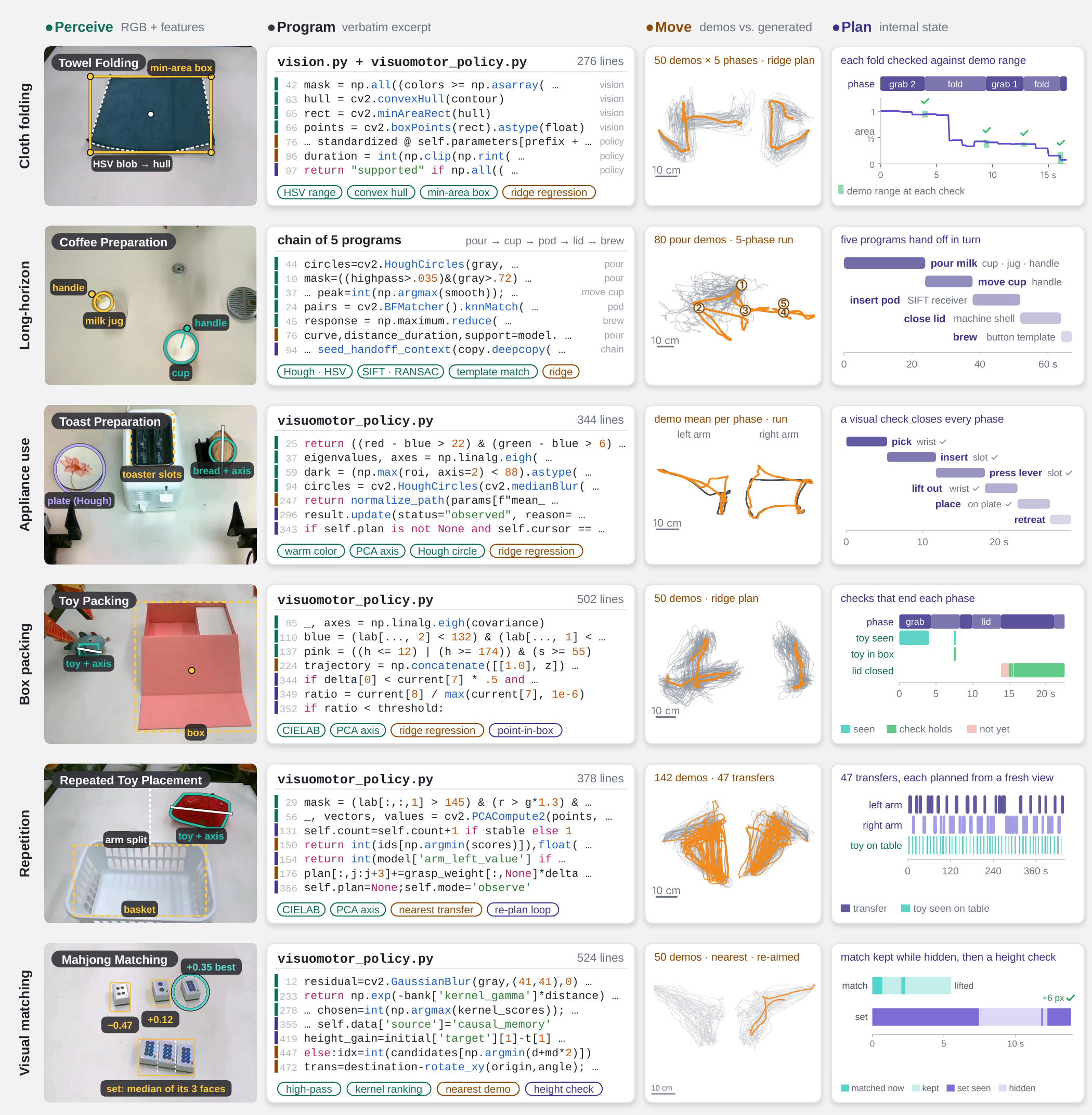}
    \caption{\small \textbf{Additional real-world program analyses (2/2).} Generated trajectories are logged commands for Coffee Preparation and Toast Preparation, and planned or replayed motions for the other tasks. Toast Preparation shows mean demonstrated motions, with each arm in its own base frame. Mahjong Matching is reconstructed offline from recorded video.}
    \label{fig:program_analysis_supp2}
\end{figure}

\Cref{fig:program_analysis_supp1,fig:program_analysis_supp2} show the remaining eleven real-world programs.

\clearpage
\section{Complete RoboTwin Simulation Results}
\label{app:sim_results}

\Cref{tab:sim_full} reports the Clean-to-Clean results on all 50 RoboTwin tasks. Baseline results are from the official RoboTwin 2.0 leaderboard~\cite{robotwin_leaderboard} (updated November 20, 2026). DP, ACT, and DP3 are trained per task, whereas $\pi_{0.5}$ is co-trained on all 50 tasks~\cite{dp,act,ze20243d,pi05}. DP3 also uses depth. The released DP, ACT, and $\pi_{0.5}$ adapters use head and wrist RGB observations and joint-space actions, whereas our programs use head-camera RGB and end-effector targets executed by the native planner~\cite{robotwin_baseline_code,xpolicylab_code}.

\begin{table}[H]
    \caption{\small
    \textbf{Complete RoboTwin 2.0 Clean-to-Clean results.}
    Each of the 50 tasks uses 100 expert-feasible test scenes from the standard random seeds.
    $^{\dagger}$ Uses depth; $^{\ddagger}$ co-trained on all 50 tasks.
    }
    \label{tab:sim_full}
    \centering
    \begingroup
    \setlength{\tabcolsep}{1.5pt}
    \renewcommand{\arraystretch}{1.32}
    \footnotesize
    \begin{tabular*}{\textwidth}{@{\extracolsep{\fill}}lrrrrr@{\hspace{8pt}}lrrrrr@{}}
        \toprule
        Task & DP & ACT & DP3$^{\dagger}$ & $\pi_{0.5}^{\ddagger}$ & Ours & Task & DP & ACT & DP3$^{\dagger}$ & $\pi_{0.5}^{\ddagger}$ & Ours \\
        \cmidrule(r){1-6}\cmidrule(l){7-12}
        Adjust Bottle & 97\% & 97\% & \textbf{99\%} & \textbf{99\%} & 98\% & Place Can Basket & 18\% & 1\% & \textbf{67\%} & 56\% & 62\% \\
        Beat Block Hammer & 42\% & 56\% & 72\% & \textbf{93\%} & \textbf{93\%} & Place Cans Plasticbox & 40\% & 16\% & 48\% & 31\% & \textbf{95\%} \\
        Blocks Ranking RGB & 0\% & 1\% & 3\% & 72\% & \textbf{100\%} & Place Container Plate & 41\% & 72\% & 86\% & 92\% & \textbf{97\%} \\
        Blocks Ranking Size & 1\% & 0\% & 2\% & 45\% & \textbf{93\%} & Place Dual Shoes & 8\% & 9\% & 13\% & \textbf{70\%} & 32\% \\
        Click Alarmclock & 61\% & 32\% & 77\% & 65\% & \textbf{83\%} & Place Empty Cup & 37\% & 61\% & 65\% & \textbf{96\%} & 83\% \\
        Click Bell & 54\% & 58\% & \textbf{90\%} & 31\% & \textbf{90\%} & Place Fan & 3\% & 1\% & 36\% & \textbf{66\%} & 49\% \\
        Dump Bin Bigbin & 49\% & 68\% & 85\% & \textbf{92\%} & 90\% & Place Mouse Pad & 0\% & 0\% & 4\% & 34\% & \textbf{80\%} \\
        Grab Roller & 98\% & 94\% & 98\% & \textbf{99\%} & 95\% & Place Object Basket & 15\% & 15\% & 65\% & 70\% & \textbf{72\%} \\
        Handover Block & 10\% & 42\% & 70\% & 30\% & \textbf{93\%} & Place Object Scale & 1\% & 0\% & 15\% & 78\% & \textbf{79\%} \\
        Handover Mic & 53\% & 85\% & \textbf{100\%} & 98\% & 99\% & Place Object Stand & 22\% & 1\% & 60\% & \textbf{85\%} & \textbf{85\%} \\
        Hanging Mug & 8\% & 7\% & 17\% & 17\% & \textbf{38\%} & Place Phone Stand & 13\% & 2\% & 44\% & 65\% & \textbf{84\%} \\
        Lift Pot & 39\% & 88\% & 97\% & \textbf{98\%} & \textbf{98\%} & Place Shoe & 23\% & 5\% & 58\% & \textbf{82\%} & 71\% \\
        Move Can Pot & 39\% & 22\% & \textbf{70\%} & 60\% & 64\% & Press Stapler & 6\% & 31\% & 69\% & \textbf{79\%} & \textbf{79\%} \\
        Move Pillbottle Pad & 1\% & 0\% & 41\% & 46\% & \textbf{68\%} & Put Bottles Dustbin & 22\% & 27\% & 60\% & 78\% & \textbf{83\%} \\
        Move Playingcard Away & 47\% & 36\% & 68\% & 84\% & \textbf{99\%} & Put Object Cabinet & 42\% & 15\% & \textbf{72\%} & 39\% & 31\% \\
        Move Stapler Pad & 1\% & 0\% & 12\% & 20\% & \textbf{72\%} & Rotate QRcode & 13\% & 1\% & 74\% & \textbf{90\%} & 76\% \\
        Open Laptop & 49\% & 56\% & 82\% & \textbf{93\%} & 81\% & Scan Object & 9\% & 2\% & 31\% & 49\% & \textbf{64\%} \\
        Open Microwave & 5\% & \textbf{86\%} & 61\% & \textbf{86\%} & 72\% & Shake Bottle Horizontally & 59\% & 63\% & \textbf{100\%} & 99\% & 80\% \\
        Pick Diverse Bottles & 6\% & 7\% & 52\% & 66\% & \textbf{70\%} & Shake Bottle & 65\% & 74\% & 98\% & \textbf{100\%} & 99\% \\
        Pick Dual Bottles & 24\% & 31\% & 60\% & 72\% & \textbf{91\%} & Stack Blocks Three & 0\% & 0\% & 1\% & 77\% & \textbf{86\%} \\
        Place A2B Left & 2\% & 1\% & 46\% & \textbf{71\%} & 69\% & Stack Blocks Two & 7\% & 25\% & 24\% & 91\% & \textbf{97\%} \\
        Place A2B Right & 13\% & 0\% & 49\% & 68\% & \textbf{75\%} & Stack Bowls Three & 63\% & 48\% & 57\% & \textbf{81\%} & 72\% \\
        Place Bread Basket & 14\% & 6\% & 26\% & 63\% & \textbf{81\%} & Stack Bowls Two & 61\% & 82\% & 83\% & \textbf{94\%} & 80\% \\
        Place Bread Skillet & 11\% & 7\% & 19\% & \textbf{76\%} & 69\% & Stamp Seal & 2\% & 2\% & 18\% & 53\% & \textbf{63\%} \\
        Place Burger Fries & 72\% & 49\% & 72\% & 82\% & \textbf{100\%} & Turn Switch & 36\% & 5\% & 46\% & \textbf{52\%} & 32\% \\
        \cmidrule(l){7-12}
        \multicolumn{6}{l}{} & Mean (50 tasks) & 28.04\% & 29.74\% & 55.24\% & 70.66\% & \textbf{78.24\%} \\
        \bottomrule
    \end{tabular*}

    \endgroup
\end{table}

\clearpage
\section{Full Task-Agnostic Prompt}
\label{app:prompt}

The following technical interface and task-agnostic instructions describe how the coding agent writes the policies. They are presented in lightly shaded, page-breakable panels for readability, with minor editorial changes to internal names and file paths. The original instruction files remain available in the repository.

\lstdefinestyle{pcpCode}{
    basicstyle=\footnotesize\ttfamily,
    columns=fullflexible,
    keepspaces=true,
    showstringspaces=false,
    breaklines=true,
    breakatwhitespace=false,
    frame=none,
    xleftmargin=1em,
    aboveskip=4pt,
    belowskip=4pt
}

\definecolor{pcpPromptBorder}{gray}{0.73}
\definecolor{pcpPromptBackground}{gray}{0.985}
\mdfdefinestyle{pcpPromptTranscript}{
    backgroundcolor=pcpPromptBackground,
    linecolor=pcpPromptBorder,
    linewidth=0.55pt,
    innerleftmargin=12pt,
    innerrightmargin=12pt,
    innertopmargin=9pt,
    innerbottommargin=9pt,
    skipabove=6pt,
    skipbelow=8pt,
    splittopskip=9pt,
    splitbottomskip=9pt
}

\subsection{Technical Interface}
\label{app:interface_text}
\begingroup
\small
\setlength{\parindent}{0pt}
\setlength{\parskip}{0.35\baselineskip}
\begin{mdframed}[style=pcpPromptTranscript]
{\footnotesize\bfseries TECHNICAL INTERFACE}\par\smallskip
\par\medskip\noindent{\bfseries Demonstrations}\par
\noindent Read all selected episodes with pixel\_policy.demonstrations.iter\_episodes(demonstrations\_path). Each episode has an index, RGB observations and aligned robot action targets. Preserve episode boundaries and record episode counts and identifiers. RGB arrays are uint8 HWC.\par
\noindent For the RoboTwin HDF5 recordings, pair RGB at frame t with the measured end-effector pose and gripper state at t+1. Drop the last image, which has no successor target. These targets are successor measured states, not logged collection-time commands. The initial robot state must align with the first image. The generic reader also supports aligned Zarr observations and stored targets. Interpret target provenance from the dataset.\par
\par\medskip\noindent{\bfseries Runtime interface}\par
\noindent A policy directory contains policy.py, a manifest and the declared local fitted parameters.\par
\par\smallskip
{\ttfamily\footnotesize
\noindent\mbox{def load\_policy(bundle\_dir): ...}\par
\noindent\mbox{class Policy:}\par
\noindent\hspace*{1em}\mbox{def reset(self): ...}\par
\noindent\hspace*{1em}\mbox{def predict(self, observation, horizon=16): ...}\par
\noindent\hspace*{1em}\mbox{def on\_action\_issued(self, action): ...}\par
\noindent\hspace*{1em}\mbox{\# Optional JSON-safe diagnostic hook}\par
\noindent\hspace*{1em}\mbox{def get\_debug\_state(self): ...}\par
}
\smallskip
\noindent Observation.rgb is the current RGB image. Observation.robot\_state is the current measured robot state. For the reported RoboTwin setting, the RGB observation comes from the head camera and the robot state is a float32 RAW16 vector. The optional initial\_state is a first-decision bootstrap field aligned with the initial observation. Other backends may expose named RGB camera views when those views are present in both the demonstration and runtime data. The generic interface also supports historical first-frame and bootstrap-only compatibility modes.\par
\noindent The program itself maintains any task progress, historical features and issued-command history. The environment does not supply task progress or object state. Reset clears episode-specific state.\par
\par\medskip\noindent{\bfseries Action targets}\par
\noindent Return exactly horizon finite floating-point rows, each with 16 values in the following order: [left x,y,z, left qw,qx,qy,qz, left gripper, right x,y,z, right qw,qx,qy,qz, right gripper]\par
\noindent Positions are absolute end-effector coordinates in meters. Orientations are unit quaternions in wxyz order. The gripper value lies in [0,1]. The quaternions q and -q represent the same orientation. The standard horizon is 16.\par
\noindent The adapter may execute only a prefix of a predicted chunk. on\_action\_issued is called only for a target actually issued. A command accepted by the execution interface does not establish that the robot reached that target. Use measured robot state to assess attainment. Preserve the platform's execution and observation cadence.\par
\par\medskip\noindent{\bfseries Policy artifacts and offline checks}\par
\noindent Save the policy, reproducible fitting code, fitted parameters, manifest, fitting report, data and source provenance, and an implementation README. Declare the artifact hashes and use local relative paths to load fitted parameters. Keep rollout records separate from the demonstrations used to write the policy.\par
\noindent The generic offline validator checks loading, reset, declared observations, action shape and values, and issued-command callbacks using recorded observations. It does not simulate physical tracking or establish rollout success. Environment evaluation uses the completed policy through the native platform adapter.\par

\end{mdframed}
\endgroup

\subsection{Task-Agnostic Coding Instructions}
\label{app:prompt_text}
\begingroup
\small
\setlength{\parindent}{0pt}
\setlength{\parskip}{0.35\baselineskip}
\begin{mdframed}[style=pcpPromptTranscript]
{\footnotesize\bfseries CODING AGENT INSTRUCTIONS}\par\smallskip
\par\noindent{\large\bfseries Robust Task-Aware Pixel Policy from Demonstrations}\par\medskip
\noindent Build a robust, executable robot policy from the supplied demonstrations. Use a lightweight \textbf{TAMP-inspired structure} to organize reusable skills, verify meaningful outcomes, and support replanning when necessary.\par
\noindent \textbf{The primary objective is reliable task completion across diverse manipulation tasks.} Autonomously discover task goals, robust visual features, motion constraints, and suitable controllers from demonstration evidence.\par
\noindent The final policy must be interpretable, non-neural Python code, without pretrained perception models, neural policies, or manually provided task solutions.\par
\noindent Prioritize \textbf{robust feature discovery, accurate complete motions, and stable execution} over architectural complexity.\par
\par\medskip\noindent{\bfseries Inputs \& Objective}\par
\begin{itemize}\setlength{\itemsep}{1pt}\setlength{\topsep}{3pt}
\item \textbf{Demonstrations:} \texttt{\detokenize{<path-to-demonstrations>}}
\item \textbf{Interface:} See the accompanying technical interface.
\item \textbf{Output:} \texttt{\detokenize{<output-directory>}}
\end{itemize}
\noindent Read the complete specified dataset using \texttt{\detokenize{pixel_policy.demonstrations.iter_episodes(demonstrations_path)}}. Verify and record its episode count and identifiers. Do not silently substitute or subsample demonstrations.\par
\noindent At every policy decision, predict the next 16 RAW16 robot actions:\par
\noindent \texttt{\detokenize{policy(current_rgb, measured_robot_state) -> actions [16, 16]}}\par
\noindent The policy maintains its history internally from earlier observations and issued commands; it is not supplied by the environment.\par
\noindent Each action contains xyz position, unit wxyz quaternion, and gripper command for both robot arms. Preserve the existing execution and observation cadence.\par
\noindent \textbf{Information boundary:} While writing and fitting the policy, use only the specified demonstrations and generic interface/adapter code. Do not inspect existing task policies, simulator task/expert implementations, privileged task metadata, task descriptions, or previous rollout results. Rollout evaluation requires separate authorization and a frozen candidate.\par
\par\medskip\noindent{\bfseries 1. Understand the Task \& Discover Skills}\par
\noindent Analyze the demonstrations to infer:\par
\begin{itemize}\setlength{\itemsep}{1pt}\setlength{\topsep}{3pt}
\item \textbf{Task goals:} Relevant entities, identities, positions, orientations, spatial relationships, and desired outcomes.
\item \textbf{Execution constraints:} Grasp consistency, contact geometry, motion ordering, bimanual coordination, and relationships to preserve.
\item \textbf{Avoidance constraints:} Undesirable contacts or object movements.
\item \textbf{Subskills:} Preconditions, subgoals, expected effects, completion evidence, and dependencies.
\item \textbf{Precision-critical intervals:} Motions where small position, orientation, contact, or gripper-timing errors may affect success.
\end{itemize}
\noindent Express important goals and constraints as measurable geometric or state-based relationships when supported by the data. Distinguish necessary requirements from incidental patterns and uncertain hypotheses.\par
\noindent \textbf{Use task understanding to guide feature selection, action fitting, and trajectory accuracy}, rather than turning approximate constraints into strict runtime gates.\par
\noindent Segment and align corresponding skills across demonstrations using RGB, robot states, actions, and interaction events. Preserve variable durations and distinct motion modes. Do not assume fixed objects, skill counts, arm roles, or task-specific primitives.\par
\noindent Before fitting, audit observation/action alignment, coordinate frames, units, quaternion ordering, gripper semantics, and recorded-command versus measured-pose meanings.\par
\par\medskip\noindent{\bfseries 2. Discover, Compare \& Validate Robust Visual Features}\par
\noindent \textbf{Robust feature discovery is a primary requirement, not an implementation detail.}\par
\noindent Autonomously identify explicit, interpretable features that reliably support the task's actual decisions: object identification, correspondence, target selection, geometric control, interaction monitoring, and outcome verification.\par
\noindent Explore appropriate classical vision and numerical methods, for example:\par
\begin{itemize}\setlength{\itemsep}{1pt}\setlength{\topsep}{3pt}
\item \textbf{Appearance:} HSV/Lab, color distributions, template matching, normalized descriptors, shape descriptors.
\item \textbf{Geometry:} Contours, edges, PCA, corners, keypoints, dimensions, orientation, relative positions.
\item \textbf{Correspondence:} SIFT/ORB, local matching, affine/homography transforms, RANSAC, fitted cross-view mappings.
\item \textbf{Temporal reasoning:} Optical flow, tracking, motion consistency, historical geometry.
\item \textbf{Visual-to-robot geometry:} Relative pose estimation, coordinate transformations, calibration, fitted geometric mappings.
\end{itemize}
\noindent These are suggestions, not mandatory algorithms. Select or combine methods based on the specific demonstrations rather than applying the same feature extractor to every task.\par
\par\smallskip\noindent{\bfseries\itshape Cross-Demonstration Feature Reliability}\par
\noindent \textbf{Do not accept a feature merely because it works on a few selected images.}\par
\noindent Systematically compare its behavior across multiple demonstrations and the relevant skill phases. Validate:\par
\begin{itemize}\setlength{\itemsep}{1pt}\setlength{\topsep}{3pt}
\item \textbf{Detection stability:} Does it reliably identify the intended feature across demonstrated configurations, viewpoints, appearances, and occlusions?
\item \textbf{Semantic correctness:} Does it distinguish the correct object, target, or correspondence rather than merely detecting something visually similar?
\item \textbf{Geometric accuracy:} Are the measured positions, orientations, and relationships sufficiently precise for the actions that depend on them?
\item \textbf{Temporal consistency:} Does the feature behave consistently across observations and interactions, accounting for camera and robot motion?
\item \textbf{Confidence reliability:} Can the policy recognize ambiguous, incorrect, or missing detections rather than confidently accepting them?
\end{itemize}
\noindent Evaluate features according to the \textbf{decisions and actions they support}, not just detection availability. Compare alternatives using held-out demonstrations and relevant intermediate frames.\par
\noindent Prefer a smaller set of reliable, discriminative features over a larger set of weak or unstable measurements. If a feature is unreliable, improve or replace its representation before compensating with arbitrary control heuristics.\par
\par\smallskip\noindent{\bfseries\itshape Camera Usage \& Causal Feature Memory}\par
\noindent Use only camera streams available in both the demonstrations and runtime interface.\par
\noindent Head cameras may provide global geometry. Wrist cameras may assist with local alignment, grasp/contact states, and interaction verification. Account for wrist-camera motion when interpreting visual changes. Wrist-based action correction is optional and must be supported by evidence.\par
\noindent Maintain a compact shared \textbf{causal Feature State} containing current features, reliable historical geometry, measured robot state, and execution history.\par
\noindent \textbf{Feature updates must be confidence-aware and geometrically consistent:}\par
\begin{itemize}\setlength{\itemsep}{1pt}\setlength{\topsep}{3pt}
\item Update estimates using new observations only when their identity, geometry, and reliability are sufficiently supported.
\item Preserve previously reliable estimates through temporary occlusion or ambiguous observations.
\item Correct or invalidate old estimates when stronger evidence contradicts them or interactions may have changed their validity.
\item Track source, observation age, confidence, and validity where relevant; never represent stale geometry as a fresh detection.
\item Avoid large or inconsistent motion corrections caused by uncertain feature updates.
\end{itemize}
\noindent Confidence and update rules should be calibrated against demonstration evidence rather than arbitrary image scores. Do not permanently freeze historical estimates when reliable new observations become available.\par
\noindent Use the same feature extraction and update logic during fitting and deployment, respecting the actual observation cadence. Never use future observations to construct earlier Feature States.\par
\par\medskip\noindent{\bfseries 3. Learn Accurate, Stable \& Reusable Skill Controllers}\par
\noindent For each skill, choose an appropriate action-generation method based on the demonstrations:\par
\begin{itemize}\setlength{\itemsep}{1pt}\setlength{\topsep}{3pt}
\item Retrieval and geometric adaptation.
\item Regression to actions or motion parameters.
\item Parameterized trajectories or movement primitives.
\item Visual servoing or feedback control.
\item Hybrid approaches.
\end{itemize}
\noindent \textbf{Do not force a particular method.} Prefer simple, stable, interpretable controllers supported by cross-demonstration validation.\par
\noindent Fit each skill using its own \textbf{causal entry Feature State} and aligned action segment. Reliable earlier observations may contribute through memory, but later skills must not be fitted solely from episode-initial features.\par
\noindent Generate accurate complete trajectories while preserving:\par
\begin{itemize}\setlength{\itemsep}{1pt}\setlength{\topsep}{3pt}
\item Position and orientation throughout motion, especially precision-critical intervals.
\item Object-relative geometry, grasp offsets, contact relationships, and motion direction.
\item Arm roles, bimanual coordination, motion phases, and gripper timing.
\item Execution/avoidance constraints and cross-skill physical consistency.
\end{itemize}
\noindent When an object remains grasped across skills, maintain its object-to-gripper relationship. Switching references or controllers must preserve compatible interaction geometry.\par
\noindent \textbf{Validate complete trajectories on held-out demonstrations, not just endpoints.} Check intermediate position/orientation errors, interaction geometry, numerical conditioning, extrapolation, and demonstrated motion support.\par
\noindent Finite outputs, smooth trajectories, and accurate endpoints alone do not establish a reliable controller. Prefer improving geometry and action fitting over masking errors with loose completion thresholds or arbitrary clipping.\par
\par\smallskip\noindent{\bfseries\itshape Stable Skill Execution}\par
\noindent \textbf{Prioritize reliable skill-entry planning and consistent execution over frequent online corrections.}\par
\noindent Each skill should normally:\par
\begin{enumerate}\setlength{\itemsep}{1pt}\setlength{\topsep}{3pt}
\item Initialize from the current reliable Feature State and measured robot state.
\item Generate a complete, accurate motion trajectory.
\item Execute consistently while preserving completed progress.
\item Verify meaningful outcomes near appropriate skill boundaries.
\end{enumerate}
\noindent \textbf{A planned motion segment may execute without visual correction at every action.}\par
\noindent Use measured robot state and actual issued-command history to estimate execution progress and maintain trajectory continuity. Do not confuse issued actions with measured attainment.\par
\noindent Do not repeatedly regenerate trajectories because of minor tracking deviations or unreliable visual changes.\par
\noindent Adjust remaining motion only when new evidence is sufficiently reliable, task-relevant, and the correction is supported by the demonstrated behavior. Continuous feedback is allowed when clearly beneficial and validated.\par
\noindent Distinguish continuing unfinished motion from restarting a skill. Restart only when the current state supports a valid new invocation; do not automatically reset progress because of execution error.\par
\par\medskip\noindent{\bfseries 4. Reliable Execution \& Lightweight TAMP}\par
\noindent Use TAMP primarily for \textbf{skill composition, task-aware verification, and occasional supported replanning}.\par
\noindent At each policy decision:\par
\begin{enumerate}\setlength{\itemsep}{1pt}\setlength{\topsep}{3pt}
\item Update relevant Feature State using current observations and measured robot state.
\item Maintain the active skill and estimate its execution progress.
\item Generate the next 16 actions from the planned motion, applying only justified corrections.
\item At meaningful skill boundaries, verify expected effects and select the next applicable skill.
\item Reconsider the plan earlier only when reliable evidence indicates significant failure, invalid applicability, or a safety concern.
\end{enumerate}
\par\smallskip\noindent{\bfseries\itshape Task-Aware Verification}\par
\noindent Distinguish \textbf{trajectory progress, measured robot attainment, and actual task effects}.\par
\noindent Skill completion should reflect meaningful task outcomes, not merely elapsed time, gripper commands, or final wrist positions. A correct endpoint must not conceal a missed critical interaction.\par
\noindent Use task-relevant geometric relationships, reliable visual evidence, measured robot state, and causal history to assess outcomes.\par
\noindent \textbf{Avoid unnecessarily strict runtime gates.} Occlusion, noisy detections, or minor deviations should not interrupt supported nominal execution. Missing evidence remains \texttt{\detokenize{unknown}}, not automatically failure or success.\par
\noindent Use inferred task constraints primarily to improve motion and guide verification. Safety-critical constraints remain strict.\par
\par\smallskip\noindent{\bfseries\itshape Skill Reuse \& Replanning}\par
\noindent Prefer completing the current skill before replanning. When necessary, select a supported retry, earlier skill, or next skill according to current conditions and task progress.\par
\noindent Every valid reinvocation must use updated Feature State and recomputed motion. Do not impose arbitrary skill-visit limits or permanent holds that block supported progress.\par
\noindent Prevent unproductive repetition based on lack of progress, unsupported configurations, and documented execution budgets. Do not invent unsupported recovery behavior.\par
\noindent A linear skill graph is acceptable when supported by the demonstrations. Learned conditions must genuinely influence transitions without unnecessary planning complexity.\par
\noindent \textbf{TAMP must support reliable task completion, not repeatedly interrupt otherwise valid execution.}\par
\par\medskip\noindent{\bfseries 5. Implementation \& Offline Validation}\par
\noindent Follow the accompanying technical interface:\par
\par\smallskip
{\ttfamily\footnotesize
\noindent\mbox{def load\_policy(bundle\_dir): ...}\par
\noindent\mbox{class Policy:}\par
\noindent\hspace*{1em}\mbox{def reset(self): ...}\par
\noindent\hspace*{1em}\mbox{def predict(self, observation, horizon=16): ...}\par
\noindent\hspace*{1em}\mbox{def on\_action\_issued(self, action): ...}\par
\noindent\hspace*{1em}\mbox{def get\_debug\_state(self): ...}\par
}
\smallskip
\noindent Use authorized current RGB observations and measured robot state at each policy decision. Document any interface mismatch and make only necessary authorized generic changes.\par
\noindent Return finite \texttt{\detokenize{[horizon,16]}} RAW16 actions with valid xyz positions, unit wxyz quaternions, and gripper values in \texttt{\detokenize{[0,1]}}.\par
\noindent Record only actually issued commands, not unexecuted predictions. Preserve the existing adapter, low-level controller, and execution cadence. Reset episode-specific memory, features, skill progress, and cached plans.\par
\par\smallskip\noindent{\bfseries\itshape Deliverables}\par
\noindent Save \texttt{\detokenize{policy.py}}, reproducible \texttt{\detokenize{fit.py}}, fitted parameters, learned task/skill structure, \texttt{\detokenize{manifest.json}}, \texttt{\detokenize{fit_report.json}}, and an English \texttt{\detokenize{README.md}}.\par
\noindent Document inferred goals and constraints, discovered features, cross-demonstration reliability, critical motion intervals, controller fitting, execution assumptions, and known limitations.\par
\par\smallskip\noindent{\bfseries\itshape Required Offline Checks}\par
\begin{enumerate}\setlength{\itemsep}{1pt}\setlength{\topsep}{3pt}
\item \textbf{Feature robustness:} Validate detection, identity matching, geometric accuracy, temporal stability, and confidence across held-out demonstrations.
\item \textbf{Feature updates:} Test reliable updates, ambiguous detections, occlusion handling, historical estimate preservation, and correction/invalidation from contradictory evidence.
\item \textbf{Motion accuracy:} Evaluate complete trajectories, precision-critical intervals, orientations, interaction geometry, gripper timing, and cross-skill consistency.
\item \textbf{Controller stability:} Check numerical conditioning, extrapolation, demonstrated motion support, and sensitivity to relevant versus irrelevant feature changes.
\item \textbf{Execution continuity:} Test partial execution and remaining-motion generation from recorded intermediate measured states, without assuming perfect tracking.
\item \textbf{TAMP:} Verify task-effect checks, appropriate skill transitions, supported retries, and avoidance of unnecessary restarts or stopping.
\item \textbf{Interface:} Validate RAW16 outputs, reset, issued-command history, partial chunks, and terminal behavior.
\end{enumerate}
\noindent Exclude held-out demonstrations from fitting and model selection, then refit final parameters on all supplied data. Distinguish ideal tracking from measured execution; unlabelled perturbations establish numerical stability, not successful physical recovery.\par
\noindent Offline validation does not establish autonomous rollout success. Keep separately authorized rollout results outside the policy bundle.\par
\noindent \textbf{Final requirement:} Deliver a complete, fitted, executable policy---not merely a task graph or design proposal.\par
\noindent \textbf{Prioritize robust, cross-demonstration-validated visual features; accurate, stable skill trajectories; and confidence-aware geometric updates. Allow planned motion segments to execute consistently when appropriate, and use lightweight TAMP for meaningful transitions and supported replanning. The primary goal is reliable task completion, not architectural complexity.}\par

\end{mdframed}
\endgroup

\end{document}